\pdfoutput=1

\documentclass[11pt]{article}

\usepackage[final]{acl}

\usepackage{times}
\usepackage{latexsym}

\usepackage[T1]{fontenc}

\usepackage[utf8]{inputenc}

\usepackage{microtype}

\usepackage{inconsolata}

\usepackage{graphicx}

\usepackage{amsmath,amsfonts,bm}
\usepackage{algorithm}
\usepackage{algpseudocode}

\usepackage{times}
\usepackage{booktabs}
\usepackage{multirow}
\usepackage{soul}
\usepackage{pgfplots}
\usepackage{hyperref}
\usepackage{url}
\usepackage[most]{tcolorbox}
\usepackage{multicol}

\usepackage{physics}
\usepackage{verbatim}
\usepackage{amssymb}
\usepackage{pifont}
\usepackage{ulem}
\usepackage{mathrsfs}
\usepackage{simplebnf}

\useunder{\uline}{\ul}{}
\usepackage{colortbl}
\PassOptionsToPackage{prologue,dvipsnames}{xcolor}

\usepackage{adjustbox}
\usepackage{subcaption}
\usepackage{seqsplit}

\definecolor{codegreen}{rgb}{0,0.6,0}
\definecolor{codegray}{rgb}{0.5,0.5,0.5}
\definecolor{codepurple}{rgb}{0.58,0,0.82}
\definecolor{backcolour}{rgb}{0.97,0.97,0.97}
\definecolor{lightblue}{rgb}{0.93,0.95,1.0}
\definecolor{LightCyan}{rgb}{0.80, 1, 1}
\definecolor{LightPink}{rgb}{1, 0.89, 0.88}
\definecolor{tablegreen}{rgb}{0.91, 0.94, 0.97}
\usepackage{listings}

\lstdefinestyle{mystyle}{
commentstyle=\color{codegreen},
keywordstyle=\color{magenta},
numberstyle=\tiny\color{codegray},
stringstyle=\color{codepurple},
basicstyle=\ttfamily\footnotesize,
breakatwhitespace=false,
breaklines=true,
captionpos=b,
keepspaces=false,
showspaces=false,
showstringspaces=false,
showtabs=false,
tabsize=2
}

\lstdefinestyle{pythonstyle}{
language=Python,
backgroundcolor=\color{backcolour},
commentstyle=\color{codegreen},
keywordstyle=\color{magenta},
numberstyle=\ttfamily\tiny\color{codegray},
stringstyle=\color{codepurple},
basicstyle=\ttfamily\tiny,
breakatwhitespace=false,
breaklines=true,
captionpos=b,
keepspaces=true,
numbers=left,
numbersep=5pt,
showspaces=false,
showstringspaces=false,
showtabs=false,
tabsize=2,
linewidth=0.95\textwidth,
xleftmargin=0.05\textwidth,
}
\lstdefinestyle{markdownstyle}{
basicstyle=\ttfamily\tiny,
backgroundcolor=\color{backcolour},
breakindent=0\dimen0,
columns=flexible,
showspaces=false,
showstringspaces=false,
breaklines=true,
breakatwhitespace=true,
breakautoindent=true,
}

\renewcommand{\algorithmiccomment}[1]{\hfill{\textcolor{codegray}{\tiny \(\triangleright\)~#1}}\par}

\newcommand{\highlight}[2]{\tikz[baseline]{\node[fill=#1, anchor=base, minimum width=2em, inner sep=1pt, outer sep=0.5pt, rounded corners] {#2};}}

\title{How to Make Large Language Models Generate 100\% Valid Molecules?}

\author{
\textbf{Wen Tao}$^{1}$ \ \ \ \ \textbf{Jing Tang}$^{2,3}$\thanks{Corresponding Author} \ \ \ \ \textbf{Alvin Chan}$^{1}$ \ \ \ \ 
\textbf{Bryan Hooi}$^{4}$ \ \ \ \ \textbf{Baolong Bi}$^{5}$ \\ 
\textbf{Nanyun Peng}$^{6}$ \ \ \ \ \textbf{Yuansheng Liu}$^{7}$\footnotemark[1] \ \ \ \ \textbf{Yiwei Wang}$^{8}$ \\ 
$^1$Nanyang Technological University \quad $^2$HKUST(GZ) \quad $^3$HKUST \\ $^4$NUS \quad $^5$UCAS \quad $^6$UCLA \quad $^7$Hunan University \quad $^8$UC Merced \\ 
\texttt{taowen228@gmail.com, jingtang@ust.hk, yuanshengliu@hnu.edu.cn}
}

\begin{document}
\maketitle
\begin{abstract}
  Molecule generation is key to drug discovery and materials science, enabling the design of novel compounds with specific properties. Large language models (LLMs) can learn to perform a wide range of tasks from just a few examples. However, generating valid molecules using representations like SMILES is challenging for LLMs in few-shot settings. In this work, we explore how LLMs can generate 100\% valid molecules. We evaluate whether LLMs can use SELFIES, a representation where every string corresponds to a valid molecule, for valid molecule generation but find that LLMs perform worse with SELFIES than with SMILES. We then examine LLMs’ ability to correct invalid SMILES and find their capacity limited. Finally, we introduce SmiSelf, a cross-chemical language framework for invalid SMILES correction. SmiSelf converts invalid \textsc{\textbf{SMI}les} to \textsc{\textbf{SELF}ies} using grammatical rules, leveraging SELFIES' mechanisms to correct the invalid SMILES. Experiments show that SmiSelf ensures 100\% validity while preserving molecular characteristics and maintaining or even enhancing performance on other metrics. SmiSelf helps expand LLMs' practical applications in biomedicine and is compatible with all SMILES-based generative models. Code is available at \url{https://github.com/wentao228/SmiSelf}.
\end{abstract}

\section{Introduction}

\begin{figure}
  \centering
  \includegraphics[width=\linewidth]{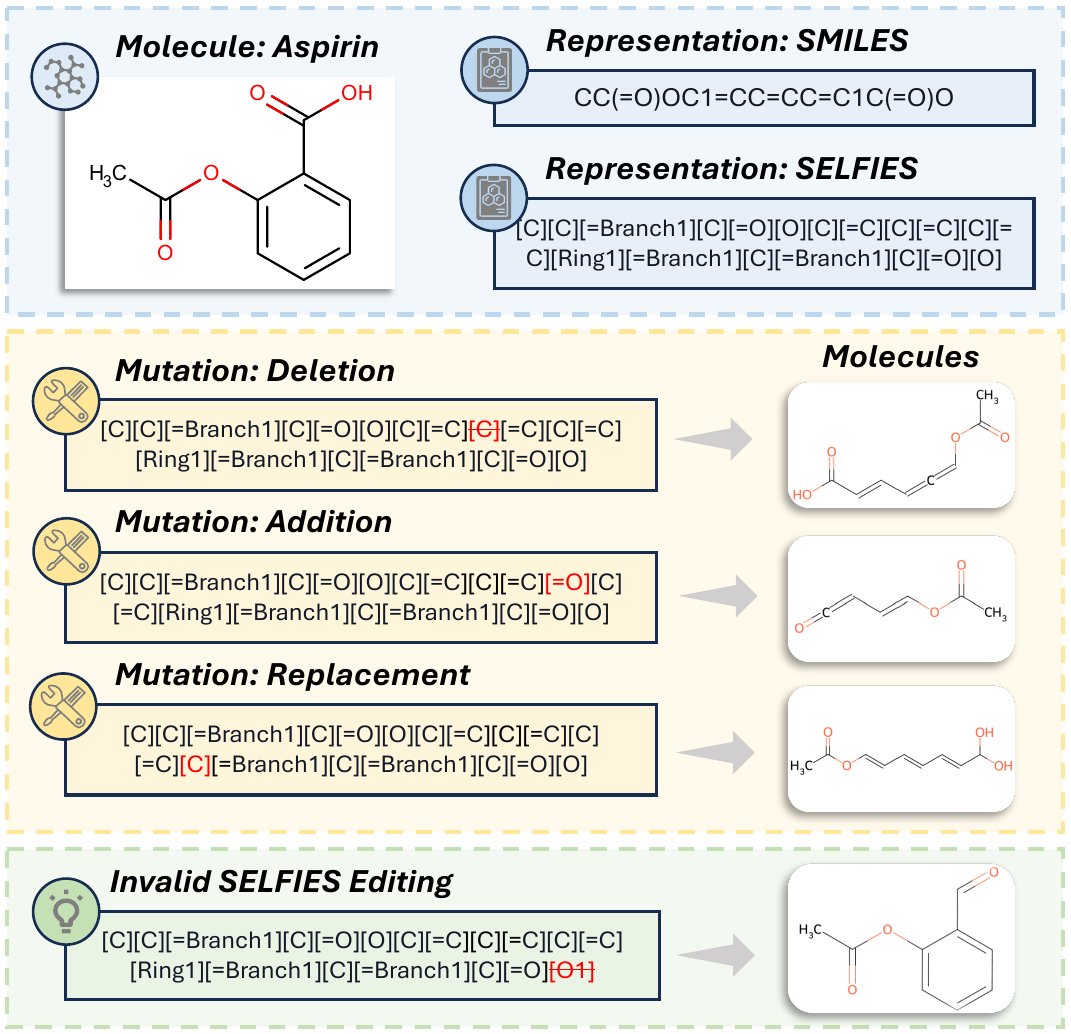}
  \caption{\textbf{Top:} SMILES and SELFIES representations of a molecule. \textbf{Middle:} Mutating the SELFIES of the molecule always results in valid molecules. \textbf{Bottom:} Proposed Invalid SELFIES Editing.}
  \vspace{-4mm}
  \label{fig:teaser}
\end{figure}

Generating novel molecules has been a fundamental and crucial problem in drug discovery and material design \cite{cheng2021molecular}. Advances in machine learning, particularly deep learning, have accelerated progress in this area \cite{zeng2022deep}. Molecules can be represented as Simplified Molecular-Input Line-Entry System (SMILES) strings \cite{weininger1988smiles} or SELF-referencIng Embedded Strings (SELFIES) \cite{krenn2020self}, both of which are compatible with language models, as shown in Figure~\ref{fig:teaser} (Top).

\begin{figure}[t]
  \centering
  \begin{tcolorbox}[fonttitle = \small\bfseries, title=Text-based molecule generation via LLMs,colframe=gray!2!black,colback=gray!2!white,boxrule=1pt,boxsep=2pt,left=5pt,right=5pt,fontupper=\footnotesize, halign title = flush center]
    \textbf{Prompt:} Given the caption of a molecule, predict the SMILES representation of the molecule.
    \tcbline
    \textbf{Input Description:} The molecule is a member of the class of tetralins that is tetralin substituted by methyl groups at positions 1, 1 and 6 respectively. It has a role as a metabolite. It is a member of tetralins and an ortho-fused bicyclic hydrocarbon. It derives from a hydride of a tetralin.
    \tcbline
    \textbf{Output SMILES:} CC1=CC2=C(C=C1)C(CCC2)(C)C\textbf{\color{red}{)}}
  \end{tcolorbox}
  \caption{An example of an invalid SMILES string generated by text-based molecule generation via LLMs.}
  \vspace{-4mm}
  \label{fig:text-based-molecule-generation}
\end{figure}

\begin{figure*}[t]
  \centering
  \includegraphics[width=\textwidth]{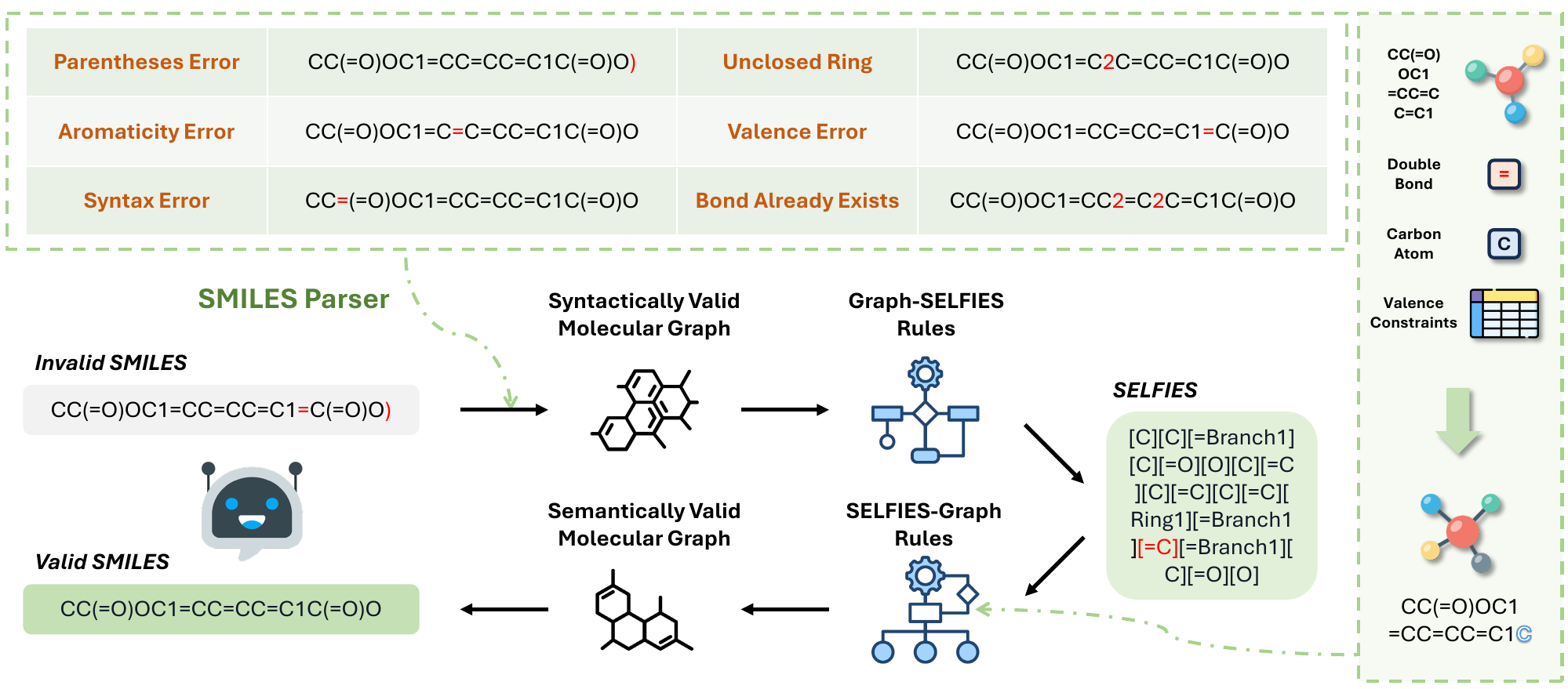}
  \vspace{-4mm}
  \caption{Overview of SmiSelf. An invalid SMILES string generated by an LLM is processed by a SMILES parser capable of handling various errors, converted into a syntactically valid molecular graph, and then transformed into a SELFIES string. The SELFIES string is re-converted into a semantically valid molecular graph, ensuring compliance with syntactic and semantic constraints, thus guaranteeing 100\% validity. Finally, the molecular graph is translated back into a valid SMILES string.}
  \vspace{-4mm}
  \label{fig:framework}
\end{figure*}

Prompting Large Language Models (LLMs), such as ChatGPT, with demonstrations has showcased their impressive ability to leverage extensive pretraining to perform diverse tasks \cite{mei2025surveycontextengineeringlarge}, opening up new opportunities for efficient and effective molecule generation. For instance, as shown in Figure~\ref{fig:text-based-molecule-generation}, given a molecule caption (i.e., a text description of a molecule's structure and properties), LLMs can generate the corresponding molecule. This enables more comprehensive and fine-grained control over molecule design. However, generating valid molecules using representations like SMILES is challenging due to strict syntax rules, such as the correct use of parentheses for branching, proper ring closure numbers, and adherence to atom valence limits. These constraints are difficult to convey through a few examples. For instance, as shown in Figure~\ref{fig:text-based-molecule-generation}, the generated SMILES string is syntactically invalid due to an extra closing parenthesis. This makes it impossible for the string to be decoded into a valid chemical structure. In contrast, SELFIES is a robust molecular representation that guarantees 100\% validity, even for randomly generated strings.

In this paper, we explore and answer three questions through extensive experiments:

\begin{itemize}
  \item \textbf{\textit{Can LLMs use SELFIES to guarantee valid molecule generation?}}\textbf{ — Yes, but at the cost of poor performance on other metrics.} Based on the robustness of SELFIES, we emphasize that SELFIES serves as a structured representation. As shown in Figure~\ref{fig:teaser} (Middle), deleting, adding, or replacing symbols still yields a valid molecule. Leveraging this property, as shown in Figure~\ref{fig:teaser} (Bottom), we propose Invalid SELFIES Editing, which directly employs SELFIES for molecular generation with LLMs, ensuring validity by filtering out non-alphabet symbols. However, we find that LLMs perform worse with SELFIES compared to SMILES, with SMILES being the most suitable representation for molecule generation using LLMs.

  \item \textbf{\textit{Can LLMs efficiently correct the invalid SMILES they generate?}}\textbf{ — No, while LLMs demonstrate potential in correcting invalid SMILES, they also face challenges in improving validity without significant degradation in other metrics.} We propose using LLMs as post-hoc invalid SMILES correctors. As shown in Algorithm~\ref{alg:main}, given invalid SMILES generated by LLMs, the model is first prompted to generate a corrected SMILES string based on the invalid SMILES and a textual description, and then uses an external tool to verify the output SMILES. The LLMs iterate this process to continuously refine the output until it becomes a valid SMILES. We find that while LLMs can correct invalid SMILES, this is accompanied by a significant reduction in other metrics, with variations in correction rates across models and error types.

  \item \textbf{\textit{How can we make LLMs generate 100\% valid molecules while keeping good performance on other metrics?}}\textbf{ — SmiSelf, a cross-chemical language framework for invalid SMILES correction.} As shown in Figure~\ref{fig:framework}, SmiSelf converts invalid SMILES generated by LLMs into SELFIES using grammatical rules, then transforms them back into SMILES, leveraging the mechanism of SELFIES to correct the invalid SMILES. Experiments demonstrate that SmiSelf guarantees 100\% validity, preserves molecular characteristics, and maintains or enhances performance on other metrics.
\end{itemize}

Overall, this work provides insights into the capabilities of current LLMs and expands their practical applications in biomedicine.

\section{LLMs Perform Worse With SELFIES}

\subsection{Molecular Representations}

As shown in Figure~\ref{fig:teaser} (Top), SMILES \cite{weininger1988smiles} and SELFIES \cite{krenn2020self} are two of the most widely used molecular representations. Like human language, the SMILES syntax enforces strict rules regarding which strings are syntactically valid. As a result, language models may generate SMILES that do not correspond to any valid chemical structure. SELFIES is a molecular string representation that guarantees 100\% robustness, ensuring that every possible combination of symbols from its alphabet corresponds to a valid chemical structure.

\subsection{Invalid SELFIES Editing}

Based on the robustness of SELFIES, we emphasize that SELFIES serves as a structured molecular representation. As shown in Figure~\ref{fig:teaser} (Middle), modifying a SELFIES string—whether by deleting a symbol, adding an alphabetic symbol, replacing a symbol, splitting the SELFIES string, or merging one SELFIES with another—always results in a valid molecule. Leveraging this property, as illustrated in Figure~\ref{fig:teaser} (Bottom), we introduce Invalid SELFIES Editing. We directly use LLMs to generate SELFIES representations for molecule generation. If the generated SELFIES are invalid (i.e., if they contain symbols not in the alphabet), we perform editing (removing non-alphabetic symbols) to make the SELFIES valid, ensuring the validity of the generated molecules.

\subsection{Task Description}

We evaluate molecular representations for LLMs using the following two tasks: \textit{Text-Based Molecule Generation} and \textit{Molecule Captioning} \cite{edwards2022translation}. The aim of \textit{Text-Based Molecule Generation} is to generate molecules that match the given natural language text describing a molecule's structures and properties. \noindent\textit{Molecule Captioning} is the reverse of text-based molecule generation, aiming to generate textual descriptions for a given molecule. Compared to the typical de novo molecule generation task \cite{polykovskiy2020molecular}, which aims to generate a variety of possible new molecules, these two tasks are much more challenging for deep generative models. They can assess the model's ability to understand molecules and generate them from descriptions.

\subsection{Experiment Setting}

We use the ChEBI-20 dataset \cite{edwards2021text2mol} and evaluation metrics identical to those used in MolT5 \cite{edwards2022translation}. The baselines include RNN \cite{cho2014learning}, Transformer \cite{vaswani2017attention}, T5 \cite{raffel2020exploring}, MolT5 \cite{edwards2022translation}, GPT-3.5, GPT-4 \cite{achiam2023gpt}, and LLaMA2 \cite{touvron2023llama}. See Appendix~\ref{appendix:molecule-caption-generation} for details.

\subsection{Experiment Results}

\begin{figure}[!ht]
  \centering
  \includegraphics[width=\linewidth]{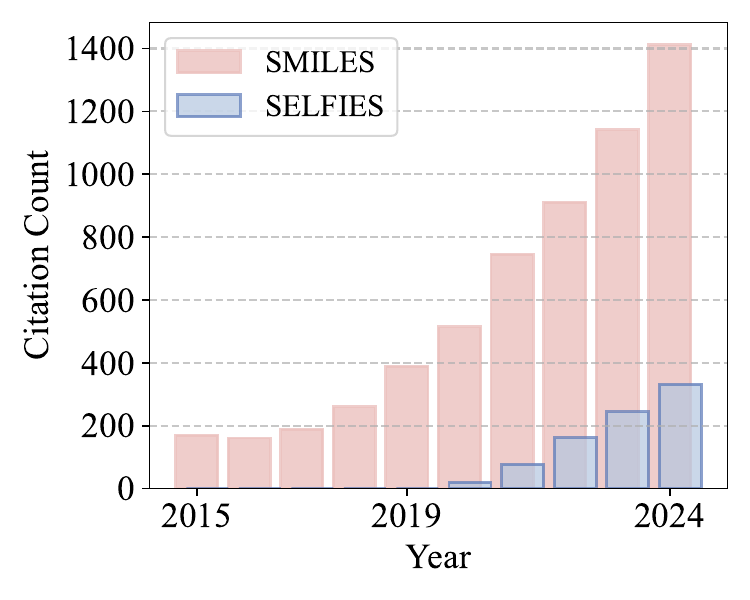}
  \vspace{-6mm}
  \caption{Comparison of Citation Counts.}
  \vspace{-4mm}
  \label{fig:citation-comparison}
\end{figure}

As shown in Tables~\ref{tab:c2m_main} and \ref{tab:m2c_main}, experimental results for both tasks indicate that LLMs perform worse when using SELFIES as a molecular representation compared to SMILES. One reason for this is that SMILES was introduced much earlier than SELFIES, resulting in its much greater presence in the training data for LLMs. Evidence supporting this can be drawn from three key aspects: First, the zero-shot results of GPT-3.5 and LLaMA2-7B in text-based molecule generation demonstrate that SMILES strings are included in their pre-training corpus, as they can generate mostly valid SMILES representations of molecules based on zero-shot prompts. Second, the zero-shot performance of GPT-3.5 and LLaMA2-7B is lower compared to task-specific small-scale models, and significantly inferior to that of T5 and MolT5 in text-based molecule generation. This suggests that these LLMs have not been specifically trained on the ChEBI-20 dataset. Finally, as shown in Figure~\ref{fig:citation-comparison}, citation counts over the past decade reveal that publications referencing SMILES substantially outnumber those mentioning SELFIES.

\begin{table*}[htb]
  \centering
  \resizebox{2.0\columnwidth}{!}{
  \begin{tabular}{c|c|c|c|c|c|c|c|c|c|c}
    \toprule
    Method & \#Params. & BLEU$\uparrow$ & EM$\uparrow$ & Levenshtein$\downarrow$ & MACCS FTS$\uparrow$ & RDK FTS$\uparrow$ & Morgan FTS$\uparrow$ & FCD$\downarrow$ & Text2Mol$\uparrow$ & Validity$\uparrow$ \\
    \midrule
    RNN (task-specific) & 56M & 0.652 & 0.005 & 38.09 & 0.591 & 0.400 & 0.362 & 4.55 & 0.409 & 0.542 \\
    Transformer (task-specific) & 76M & 0.499 & 0.000 & 57.66 & 0.480 & 0.320 & 0.217 & 11.32 & 0.277 & 0.906 \\
    \midrule
    T5-Base (fine-tuned) & 248M & 0.762 & 0.069 & 24.950 & 0.731 & 0.605 & 0.545 & 2.48 & 0.499 & 0.660 \\
    T5-Large (fine-tuned) & 783M & \underline{0.854} & 0.279 & \underline{16.721} & 0.823 & 0.731 & 0.670 & 1.22 & 0.552 & 0.902 \\
    MolT5-Base & 248M & 0.769 & 0.081 & 24.458 & 0.721 & 0.588 & 0.529 & 2.18 & 0.496 & 0.772 \\
    MolT5-Large & 783M & \underline{0.854} & \textbf{0.311} & \textbf{16.071} & 0.834 & \underline{0.746} & \underline{0.684} & 1.20 & 0.554 & \underline{0.905} \\
    \midrule
    LLaMA2-7B (zero-shot) & 7B & 0.104 & 0.000 & 84.18 & 0.243 & 0.119 & 0.089 & 42.01 & 0.148 & 0.631\\
    LLaMA2-7B (2-shot) & 7B & 0.693 & 0.022 & 36.77 & 0.808 & 0.717 & 0.609 & 4.90 & 0.149 & 0.761 \\
    GPT-3.5 (zero-shot) & N/A & 0.489 & 0.019 & 52.13 & 0.705 & 0.462 & 0.367 & 2.05 & 0.479  & 0.802 \\
    GPT-3.5 (10-shot) & N/A & 0.790 & 0.139 & 24.91 & \underline{0.847} & 0.708 & 0.624 & \underline{0.57} & \underline{0.571} & 0.887 \\
    \midrule
    GPT-4 (10-shot) & 1.76T & \textbf{0.857} & \underline{0.280} & 17.14 & \textbf{0.903} & \textbf{0.805} & \textbf{0.739} & \textbf{0.41} & \textbf{0.593} & 0.899 \\
    \rowcolor[HTML]{EFEFEF}
    GPT-4-SELFIES (10-shot) & 1.76T & 0.682 & 0.179 & 26.596 & 0.756 & 0.624 & 0.541 & 1.666 & 0.468 & \textbf{1.000} \\
    \bottomrule
  \end{tabular}
  }
  \caption{Text-based molecule generation results on ChEBI-20. The \textbf{best} scores are in bold, and the \underline{second-best} scores are underlined. ``N/A'' indicates that the parameter size is unknown.}
  \label{tab:c2m_main}

\end{table*}

\begin{table*}[htb]
  \centering
  \resizebox{2.0\columnwidth}{!}{
  \begin{tabular}{c|c|c|c|c|c|c|c|c}
    \toprule
    Methods & \#Params. & BLEU-2$\uparrow$ & BLEU-4$\uparrow$ & ROUGE-1$\uparrow$ & ROUGE-2$\uparrow$ & ROUGE-L$\uparrow$ & METEOR$\uparrow$ & Text2Mol$\uparrow$ \\
    \midrule
    RNN (task-specific) & 56M & 0.251 & 0.176 & 0.450 & 0.278 & 0.394 & 0.363 & 0.426 \\
    Transformer (task-specific) & 76M & 0.061 & 0.027 & 0.204 & 0.087 & 0.186 & 0.114 & 0.057 \\
    \midrule
    T5-Base (fine-tuned) & 248M & 0.511 & 0.423 & 0.607 & 0.451 & 0.550 & 0.539 & 0.523 \\
    T5-Large (fine-tuned) & 783M & 0.558 & 0.467 & 0.630 & 0.478 & 0.569 & 0.586 & 0.563 \\
    MolT5-Base & 248M & 0.540 & 0.457 & \underline{0.634} & \underline{0.485} & \underline{0.578} & 0.569 & 0.547 \\
    MolT5-Large & 783M & \underline{0.594} & \underline{0.508} & \textbf{0.654} & \textbf{0.510} & \textbf{0.594} & \textbf{0.614} & \underline{0.582} \\
    \midrule
    LLaMA2-7B (zero-shot) & 7B & 0.094 & 0.039 & 0.169 & 0.054 & 0.142 & 0.175 & 0.153\\
    LLaMA2-7B (2-shot) & 7B & 0.489 & 0.409 & 0.535 & 0.374 & 0.472 & 0.495 & 0.466\\
    GPT-3.5 (zero-shot) & N/A & 0.103 & 0.050 & 0.261 & 0.088 & 0.204 & 0.161 & 0.352 \\
    GPT-3.5 (10-shot) & N/A & 0.565 & 0.482 & 0.623 & 0.450 & 0.543 & 0.585 & 0.560 \\
    \midrule
    GPT-4 (10-shot) & 1.76T & \textbf{0.607} & \textbf{0.525} & \underline{0.634} & 0.476 & 0.562 & \underline{0.610} & \textbf{0.585} \\
    \rowcolor[HTML]{EFEFEF}
    GPT-4-SELFIES (10-shot) & 1.76T & 0.569 & 0.488 & 0.607 & 0.445 & 0.538 & 0.577 & 0.550 \\
    \bottomrule
  \end{tabular}
  }
  \caption{The performance of molecule captioning on ChEBI-20. The \textbf{best} scores are in bold, and the \underline{second-best} scores are underlined. ``N/A'' indicates that the parameter size is unknown.}
  \vspace{-4mm}
  \label{tab:m2c_main}
\end{table*}

Another reason lies in the inherent characteristics of the representations themselves. Several studies \cite{skinnider2024invalid, skinnider2021chemical, edwards2022translation, guo2023can} have shown that language models trained on SMILES outperform those trained on SELFIES. Surprisingly, although models may produce invalid molecules using the SMILES format, a significantly larger number of SELFIES was required to train a model of equivalent quality to one trained on SMILES strings \cite{skinnider2021chemical}.

From the experimental results, we also observe that increasing the size of the language model leads to significant performance improvements. While it is well known that scaling up model size and pretraining data generally enhances performance \cite{kaplan2020scaling}, it is still surprising to see that when using SMILES as the molecular representation, LLMs outperform MolT5-Large—specifically pre-trained and fine-tuned for text-based molecule generation—across most metrics, with only 10-shot in-context examples.

In summary, while our proposed Invalid SELFIES Editing ensures the validity of generated molecules, \textbf{LLMs perform worse when using SELFIES. SMILES remains the most suitable molecular representation for molecule generation using language models.}

\section{LLMs Are Inefficient Invalid SMILES Correctors}

In Section~\ref{section:methods_validity}, we discuss approaches to address the validity issue in LLM-based SMILES generation and explain why they cannot fully resolve it by analyzing their limitations and comparing performance. Recent research \cite{zhong2024harnessing} has demonstrated that LLMs can function as post-hoc correctors, proposing corrections for tasks like molecular property prediction. This section explores the question: \textit{Can LLMs efficiently correct the invalid SMILES they generate?}

\subsection{Iterative SMILES Generation}

To answer the question, we propose using LLMs as post-hoc invalid SMILES correctors. Given an invalid SMILES string, an LLM is first prompted to generate a possibly valid SMILES string based on the current invalid SMILES and a textual description of the desired molecule. This output is then verified using the external tool RDKit \cite{landrum2013rdkit}. This process is repeated iteratively, where the cycle of ``\textit{Correct SMILES} $\Rightarrow$ \textit{Verify SMILES}'' continues until the generated SMILES string is valid. See Algorithm~\ref{alg:main} for a summary of the method.

\subsection{Experiment Setting}

We evaluate on the \textit{Text-Based Molecule Generation} task using the ChEBI-20 dataset \cite{edwards2021text2mol} and evaluation metrics identical to those used in MolT5 \cite{edwards2022translation}. The baseline results are 10-shot example results of GPT-3.5 and GPT-4 \cite{achiam2023gpt}. LLMs used as post-hoc correctors include GPT-3.5, GPT-4o-mini \cite{hurst2024gpt}, LLaMA2 \cite{touvron2023llama}, and LLaMA3 \cite{grattafiori2024llama}. See the Appendix~\ref{appendix:molecule-caption-generation} for further details.

\noindent\textbf{Types of Errors.} To assess the validity of model outputs, we used RDKit to identify invalid SMILES generated by LLMs—those that could not be converted into valid molecules. These invalid SMILES were classified into six categories based on RDKit error messages: syntax error, unclosed ring, parentheses error (extra open or close parentheses), bond already exists (dual occurrence of a bond between the same two atoms), aromaticity error (non-ring atom marked aromatic and kekulization errors), and valence error (exceeding an atom’s maximum number of bonds). If strings contain multiple error types, only the first error is reported.

\noindent\textbf{Correction Rate.} To evaluate how effectively the model can self-correct, we introduce the correction rate, which is the ratio of valid SMILES generated after correction to the total number of invalid SMILES before correction.

\begin{figure}[t]
  \begin{algorithm}[H]
    \small
    \begin{algorithmic}[1]
      \Require Input description $x$, initial invalid SMILES $\hat{y_0}$, prompt $p$, model $\mathcal{M}$, external tool $\mathcal{T}$, number of iterations $n$
      \State Get initial invalid SMILES $\hat{y_0}$ \algorithmiccomment{Initialization}
      \For{$i \leftarrow 0$ to $n-1$}
      \State $\hat{y_{i+1}} \sim \mathbb{P}_{\mathcal{M}}(\cdot|p\oplus x\oplus y_{i})$ \algorithmiccomment{Correction}
      \State Verify $\hat{y_{i+1}}$ through $\mathcal{T}$ to obtain feedback $f_i$ \algorithmiccomment{Verification}
      \If{$f_i$ indicates that $\hat{y_{i+1}}$ is valid} \algorithmiccomment{Stopping Criteria}
      \State \Return $\hat{y_{i+1}}$
      \EndIf
      \EndFor
      \State \Return $\hat{y_{n}}$
    \end{algorithmic}
    \caption{LLMs as invalid SMILES correctors}
    \label{alg:main}
  \end{algorithm}
  \vspace{-4mm}
\end{figure}

\subsection{Experiment Results}

As shown in Table~\ref{tab:self-refine}, LLMs can improve the validity of molecules generated by them with feedback from an external tool. However, this enhancement is accompanied by a reduction in other metrics. In particular, there is a noticeable reduction in both the BLEU score and the Levenshtein score, as well as a slight reduction in other metrics. These results indicate that, while the molecules are corrected to be valid, they deviate more from the ground truth and become less aligned with the given description, despite the description being provided during the refinement process.

\begin{figure}[!ht]
  \centering
  \includegraphics[width=\linewidth]{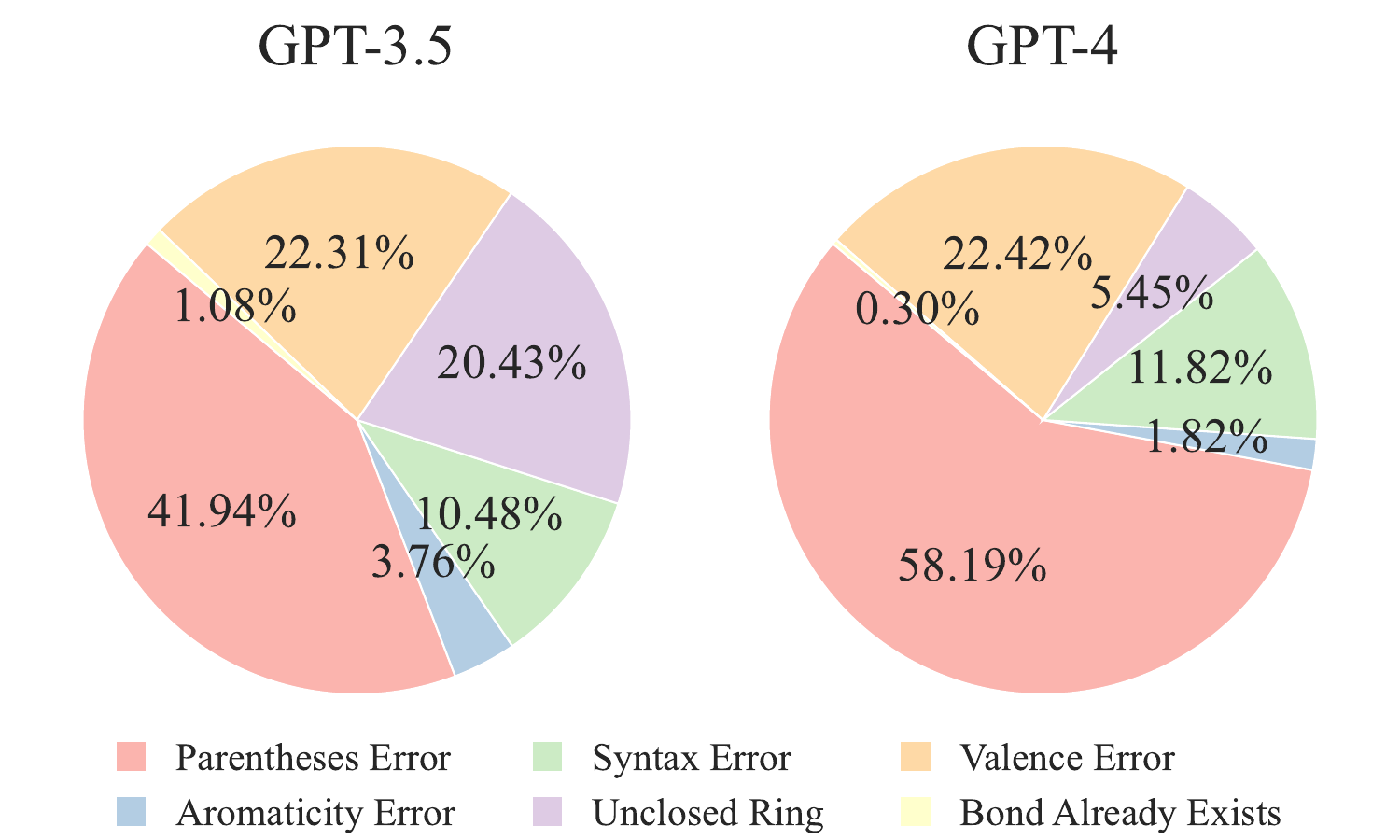}
  \caption{Distribution of error types in the invalid SMILES generated by LLMs.}
  \label{fig:Error}
\end{figure}

As shown in Figure \ref{fig:Error}, LLMs predominantly produced ``parentheses error'', which accounted for approximately half of all invalid SMILES. The second most common error was ``valence errors'', constituting 22.31\% of invalid SMILES generated by GPT-3.5 and 22.42\% by GPT-4.

\begin{figure}[!ht]
  \centering
  \begin{subfigure}[b]{0.48\textwidth}
    \centering
    \includegraphics[width=\linewidth]{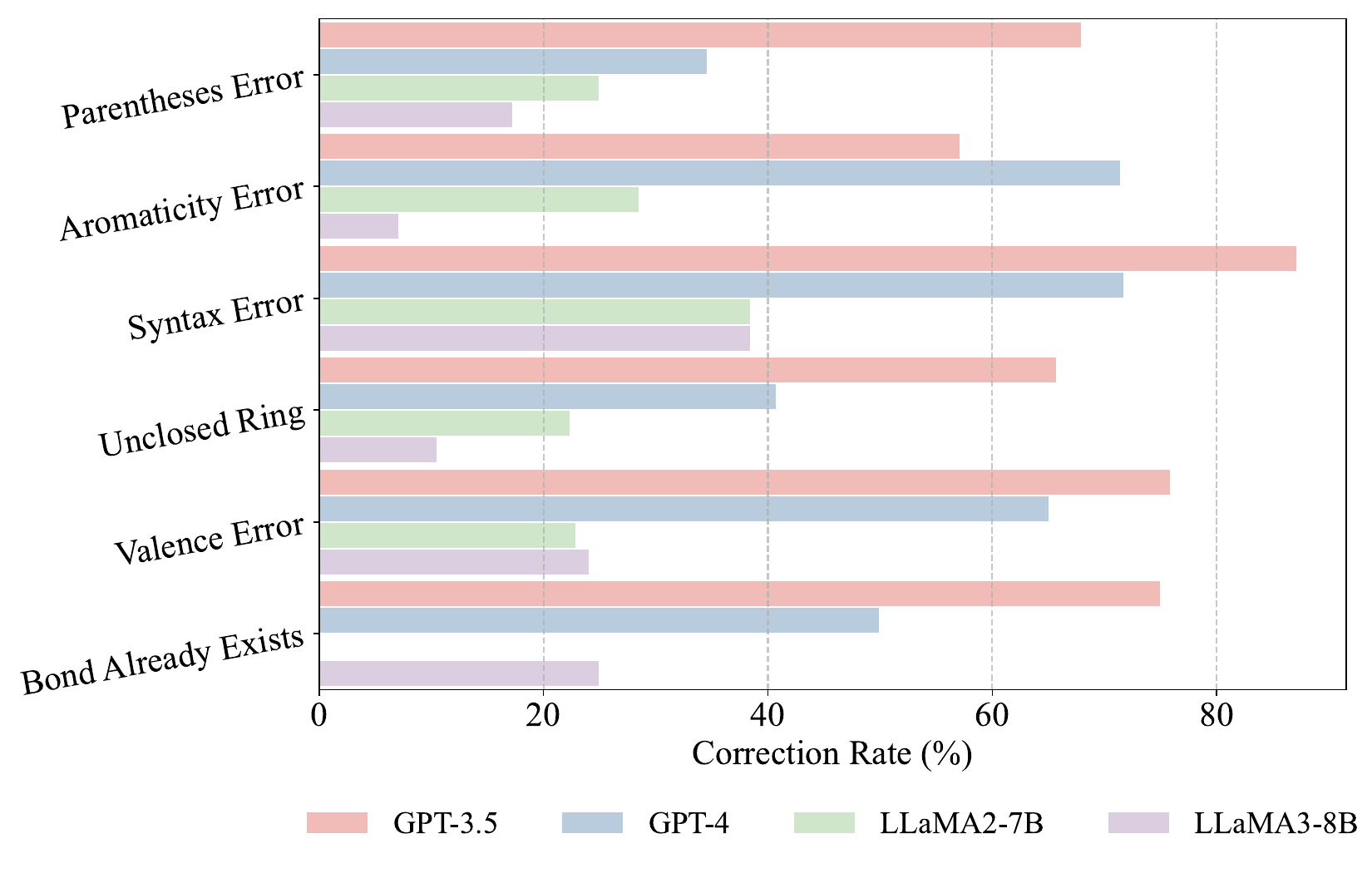}
    \caption{GPT-3.5}
    \label{fig:GPT-3.5-Turbo-Error-Correction}
  \end{subfigure}\hfill
  \begin{subfigure}[b]{0.48\textwidth}
    \centering
    \includegraphics[width=\linewidth]{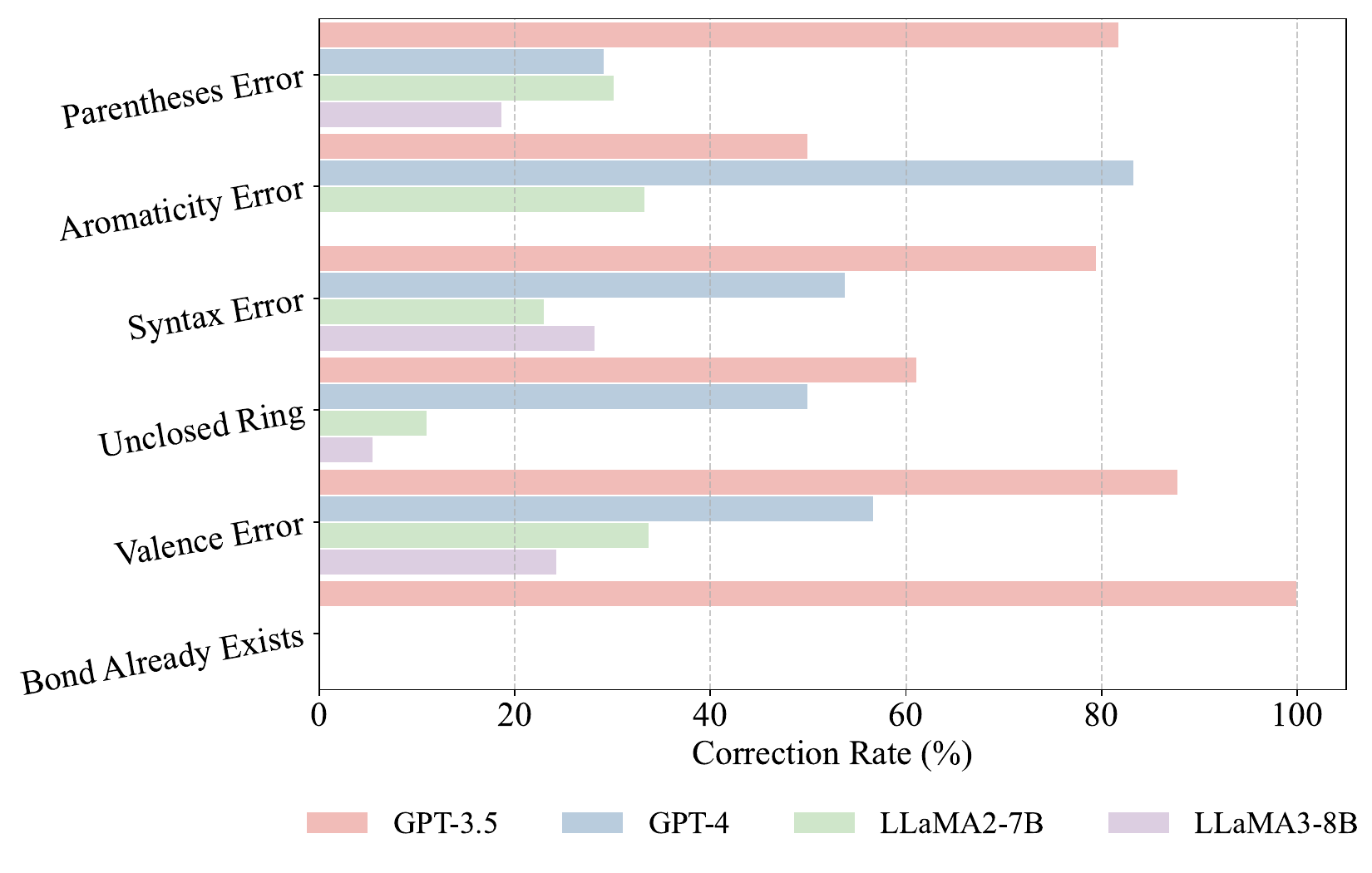}
    \caption{GPT-4}
    \label{fig:GPT-4-0314-Error-Correction}
  \end{subfigure}
  \caption{Comparison of correction rates across different LLMs for various error types in invalid SMILES generated by GPT-3.5 and GPT-4.}
  \label{fig:Error-Correction}
\end{figure}

As shown in Figure~\ref{fig:Error-Correction}, LLMs demonstrate potential in correcting invalid SMILES. However, there are significant variations in correction rates across different models and error types. Overall, the GPT series tends to outperform the LLaMA series in correcting various errors, with GPT-3.5 notably surpassing all other models.

In summary, for the text-based molecule generation task, \textbf{LLMs have demonstrated potential in correcting invalid SMILES strings. However, they continue to face challenges in enhancing validity while maintaining other metrics without significant degradation.} Additionally, there are notable variations in correction rates across different models and error types.

\begin{table*}[t]
  \centering
  \tabcolsep=4.2pt
  \resizebox{1\textwidth}{!}{
  \begin{tabular}{l|c|c|c|c|c|c|c|c|c}
    \toprule[2pt]
    Model & BLEU$\uparrow$  & EM$\uparrow$ & Levenshtein$\downarrow$ & MACCS FTS$\uparrow$ & RDK FTS$\uparrow$ & Morgan FTS$\uparrow$ & FCD$\downarrow$ & Text2Mol$\uparrow$ & Validity$\uparrow$ \\
    \midrule
    \rowcolor[HTML]{EFEFEF}
    \multicolumn{10}{c}{\textbf{GPT-3.5 (10-shot)}}\\
    \midrule
    Baseline & \textbf{0.790} & 0.139 & \textbf{24.910} & \textbf{0.847} & \textbf{0.708} & \textbf{0.624} & 0.570 & \textbf{0.571} & 0.887 \\
    + LLM Corrector (GPT-3.5)& 0.738 & \textbf{0.141} & 29.171 & 0.836 & 0.692 & 0.600 & \textbf{0.537} & 0.561 & \textbf{0.967} \\
    + LLM Corrector (GPT-4o-mini)& 0.753 & 0.140 & 29.018 & 0.837 & 0.693 & 0.606 & 0.550 & 0.563 & 0.942 \\
    + LLM Corrector (LLaMA2-7B)& 0.685 & 0.139 & 30.663 & 0.830 & 0.692 & 0.609 & 0.776 & 0.556 & 0.916 \\
    + LLM Corrector (LLaMA3-8B)& 0.732 & 0.140 & 31.400 & 0.841 & 0.700 & 0.615 & 0.568 & 0.566 & 0.909 \\
    \midrule
    \rowcolor[HTML]{EFEFEF}
    \multicolumn{10}{c}{\textbf{GPT-4 (10-shot)}}\\
    \midrule
    Baseline & \textbf{0.857} & \textbf{0.280} & \textbf{17.140} & \textbf{0.903} & \textbf{0.805} & \textbf{0.739} & 0.410 & \textbf{0.593} & 0.899 \\
    + LLM Corrector (GPT-3.5) & 0.772 & 0.280 & 22.220 & 0.890 &  0.784 & 0.710 &  \textbf{0.375} & 0.582 & \textbf{0.981} \\
    + LLM Corrector (GPT-4o-mini) & 0.788 & 0.280 & 21.299 & 0.894 &  0.790 & 0.722 &  0.401 & 0.586 & 0.940 \\
    + LLM Corrector (LLaMA2-7B)& 0.718 & 0.280 & 25.034 & 0.886 & 0.787 & 0.722 & 0.562 & 0.578 & 0.929 \\
    + LLM Corrector (LLaMA3-8B)& 0.779 & 0.280 & 24.549 & 0.897 & 0.795 & 0.728 & 0.405 & 0.587 & 0.920 \\
    \bottomrule[1.2pt]
  \end{tabular}%
  }
  \caption{Results of using LLMs as post-hoc correctors for correcting invalid SMILES in text-based molecule generation. The \textbf{best} scores are in bold.}
  \vspace{-4mm}
  \label{tab:self-refine}%
\end{table*}%

\section{Making LLMs Generate 100\% Valid Molecules}

We present SmiSelf (invalid \textsc{\textbf{SMI}les} to \textsc{\textbf{SELF}ies}), a cross-chemical language framework that ensures valid molecule generation through mutual conversion between two chemical languages: SMILES and SELFIES.

\subsection{SmiSelf: Cross-Chemical Language for Invalid SMILES Correction}

Although SELFIES is a 100\% robust molecular string representation, based on our observations, it is not as suitable as SMILES for molecule generation with LLMs. We propose converting invalid SMILES generated by LLMs into SELFIES, then transforming them back into SMILES, leveraging the mechanism of SELFIES to correct the SMILES.

SMILES and SELFIES, though both string-based molecular representations, have distinct grammars. Precise conversion that preserves molecular characteristics from invalid SMILES to SELFIES cannot be fully achieved through in-context learning. To achieve this precise conversion, we use molecular graphs as intermediaries to convert between these two representations, as molecules can be represented as molecular graphs that adhere to chemical constraints. Our goal is to eliminate both syntactic and semantic errors in invalid SMILES to ensure syntactic and semantic validity. Syntactic errors involve strings that cannot be interpreted as molecular graphs, while semantic errors involve strings that form molecular graphs but violate basic chemical rules. See the Appendix~\ref{appendix:syntactic_semantic_validity} for details of the distinction between syntactic validity and semantic validity.

As shown in Figure~\ref{fig:framework}, we implement a SMILES parser that converts invalid SMILES into a molecular graph. The string \texttt{CC(=O)OC1=CC=CC=C1=C(=O)O)} represents an invalid SMILES with both syntactic (extra closing parenthesis) and semantic (exceeding valence bond limits) errors. Carbon ``\texttt{C}'' and oxygen ``\texttt{O}'' atoms are parsed as nodes, connected by edges representing single (denoted by no symbol between atoms) or double (denoted by ``\texttt{=}'') bonds. The number ``\texttt{1}'' shows that the ring is formed between the two carbon atoms labeled ``\texttt{C1}''. Branched structures ``\texttt{(=O)}'' are given using brackets. We skip the extra closing parenthesis based on predefined rules and ignore the semantic error during graph construction, ensuring syntactic validity.

Next, the molecular graph is converted into a SELFIES string using the Graph-SELFIES rules (SELFIES grammar) \cite{krenn2020self}. The SELFIES string is transformed into a semantically valid molecular graph according to the SELFIES-Graph rules (Table~\ref{tab:SELFIES_rules}). As shown in Figure~\ref{fig:framework}, the molecule is constructed from the partial SELFIES string, corresponding to the SMILES string \texttt{CC(=O)OC1=CC=CC=C1}. The next SELFIES symbol, \texttt{[=C]}, adds a carbon atom with a double bond. However, this would violate valence constraints, so the bond is converted to a single bond by the SELFIES-Graph rules. Finally, the molecular graph is translated back into a SMILES string. These transformations ensure the resulting SMILES satisfy both syntactic and semantic constraints, guaranteeing 100\% validity.

\subsection{Experiment Setting}

\subsubsection{Text-Based Molecule Generation}

For this task, we use the ChEBI-20 dataset \cite{edwards2021text2mol} and evaluation metrics identical to those used in MolT5 \cite{edwards2022translation}. The baseline results include n-shot (0, 1, 2, 5, and 10) in-context example results of GPT-3.5 and 10-shot in-context example results of GPT-4. These are used to evaluate the performance of our proposed SmiSelf method in correcting SMILES strings generated by LLMs at varying quality levels. See the Appendix~\ref{appendix:molecule-caption-generation} for further details.

\subsubsection{Class-Specific Molecule Generation}

This task aims to generate molecules specific to a given class, based on a limited number of exemplars from that class. The dataset contains $32$ Acrylates, $11$ Chain Extenders, and $11$ Isocyanates. For each class, 100 molecules are generated using LLMs. The evaluation metrics include: Validity, percentage of chemically valid molecules, diversity, average pairwise Tanimoto distance over Morgan fingerprints \cite{rogers2010extended}; Membership, and the percentage of molecules that belong to the desired monomer class.

We employ grammar prompting \cite{wang2024grammar} during in-context learning to evaluate the benefits of explicitly incorporating generic SMILES grammar into LLM-based molecule generation. Grammar prompting enables LLMs to incorporate external knowledge and domain-specific constraints, expressed through a Backus–Naur Form grammar, during in-context learning.

Unlike prompting-based methods, the baseline model DEG \cite{guo2022data} generates molecules using a graph-based grammar, which is learned through a sequence of production rules automatically derived from the training data.
\subsection{Experiment Results}

\subsubsection{Text-Based Molecule Generation}

As shown in Table~\ref{tab:c2m_correct}, the molecules corrected by SmiSelf are 100\% valid. However, we also observe that some metrics for the corrected molecules are worse than those for the uncorrected ones. These results are to be expected. First, the calculation of metrics—excluding BLEU, Exact Match, Levenshtein, and Validity—considers only valid SMILES, so the correction phase broadens the scope of the metrics. Second, since generating molecules from molecular descriptions is a one-to-one task with a ground truth, the process of correcting the molecules inevitably introduces some distortion, which can affect the original information and slightly reduce the metrics.

These results align with those of TGM-DLM \cite{gong2024text}, which trains a diffusion model in its second phase to correct invalid SMILES generated in the first phase. However, the performance reduction observed with our method is significantly lower across most metrics compared to the reduction caused by TGM-DLM’s second phase, as well as our previously proposed Invalid SELFIES Editing and LLM Corrector. Additionally, SmiSelf improves the EM score, indicating that some invalid SMILES can exactly match the ground truth after correction, whereas TGM-DLM's EM score remains unchanged. We provide these additional comparison results in the Table~\ref{tab:validity-comparison}.

\begin{table*}[htb]
  \centering
  \resizebox{2.0\columnwidth}{!}{
  \begin{tabular}{c|c|c|c|c|c|c|c|c|c}
    \toprule
    Method & BLEU$\uparrow$ & EM$\uparrow$ & Levenshtein$\downarrow$ & MACCS FTS$\uparrow$ & RDK FTS$\uparrow$ & Morgan FTS$\uparrow$ & FCD$\downarrow$ & Text2Mol$\uparrow$ & Validity$\uparrow$ \\
    \midrule
    GPT-3.5 (zero-shot) & 0.489 & 0.019 & 52.13 & \textbf{0.705} & \textbf{0.462} & \textbf{0.367} & 2.05 & \textbf{0.479}  & 0.802 \\
    \rowcolor{tablegreen}
    \textbf{\hspace{8mm} + SmiSelf} & \textbf{0.544} & \textbf{0.020} & \textbf{46.967} & 0.688 &  0.447 & 0.339 &  \textbf{1.986} & 0.456 & \textbf{1.000} \\
    \midrule
    GPT-3.5 (1-shot) & \textbf{0.706} & 0.074 & \textbf{33.38} & \textbf{0.799} & \textbf{0.620} & \textbf{0.526} & 0.84 & \textbf{0.540} & 0.842\\
    \rowcolor{tablegreen}
    \textbf{\hspace{8mm} + SmiSelf} & 0.701 & \textbf{0.075} & 33.49 & 0.790 &  0.610 & 0.505 &  \textbf{0.719} & 0.527 & \textbf{1.000} \\
    \midrule
    GPT-3.5 (2-shot) & \textbf{0.748} & 0.101 & \textbf{28.89} & \textbf{0.827} & \textbf{0.668} & \textbf{0.578} & 0.67 & \textbf{0.557} & 0.860 \\
    \rowcolor{tablegreen}
    \textbf{\hspace{8mm} + SmiSelf} & 0.741 & \textbf{0.102} & 29.311 & 0.815 &  0.654 & 0.552 &  \textbf{0.604} & 0.545 & \textbf{1.000} \\
    \midrule
    GPT-3.5 (5-shot) & \textbf{0.771} & 0.121 & \textbf{26.78} & \textbf{0.836} & \textbf{0.686} & \textbf{0.599} & 0.60 & \textbf{0.564} & 0.882 \\
    \rowcolor{tablegreen}
    \textbf{\hspace{8mm} + SmiSelf} & 0.761 & \textbf{0.122} & 27.51 & 0.827 &  0.674 & 0.576 &  \textbf{0.542} & 0.554 & \textbf{1.000} \\
    \midrule
    GPT-3.5 (10-shot) & \textbf{0.790} & 0.139 & \textbf{24.91} & \textbf{0.847} & \textbf{0.708} & \textbf{0.624} & 0.57 & \textbf{0.571} & 0.887 \\
    \rowcolor{tablegreen}
    \textbf{\hspace{8mm} + SmiSelf} & 0.778 & \textbf{0.141} & 25.938 & 0.838 &  0.695 & 0.602 &  \textbf{0.492} & 0.561 & \textbf{1.000} \\
    \midrule
    GPT-4 (10-shot) & \textbf{0.857} & 0.280 & \textbf{17.14} & \textbf{0.903} & \textbf{0.805} & \textbf{0.739} & 0.41 & \textbf{0.593} & 0.899 \\
    \rowcolor{tablegreen}
    \textbf{\hspace{8mm} + SmiSelf} & 0.846 & \textbf{0.282} & 17.668 & 0.892 &  0.789 & 0.718 &  \textbf{0.312} & 0.584 & \textbf{1.000} \\
    \bottomrule
  \end{tabular}
  }

  \caption{Few-shot text-based molecule generation results on ChEBI-20, along with results corrected by SmiSelf. The \textbf{better} scores are in bold.}

  \label{tab:c2m_correct}

\end{table*}

\begin{table*}[htb]
  \center
  \small
  \begin{tabular}{ l c cccc c cccc c cccc  }
    \toprule
    & \,\, & \multicolumn{3}{c}{\textbf{Acrylates}} & \,\, & \multicolumn{3}{c}{\textbf{Chain Extenders}} &\,\,  &\multicolumn{3}{c}{ \textbf{Isocyanates} }\\
    Model &  & V & D & M  & & V & D & M & & V & D & M \\
    \midrule
    Graph Grammar & & 100  & 0.83  & 30  & & 100  & 0.86 & 98  & & 100  & 0.93  & 83 \\
    \midrule
    Standard Prompting  & & 23 & 0.74  &  19  & & \textbf{100}  & \textbf{0.81}   &  \textbf{99}  & & 94  & \textbf{0.82}  & 94  \\
    \rowcolor{tablegreen}
    \textbf{\hspace{8mm} + SmiSelf}  & & \textbf{100} & \textbf{0.75}  & \textbf{83}  & & \textbf{100}  & \textbf{0.81}   &  \textbf{99}  & & \textbf{100}  & \textbf{0.82}  & \textbf{100}  \\
    \midrule
    Grammar Prompting  & & 97  & 0.77  &  56  & & 86  & 0.91  & 84  &	& 71 & \textbf{0.83} & 65 \\
    \rowcolor{tablegreen}
    \textbf{\hspace{8mm} + SmiSelf}  & & \textbf{100}  & \textbf{0.78}  &  \textbf{57}  & & \textbf{100}  & \textbf{0.92}  & \textbf{96}  &	& \textbf{100} & \textbf{0.83} & \textbf{79} \\
    \bottomrule
  \end{tabular}
  \caption{Results for class-specific molecule generation with GPT-3.5, along with results corrected by SmiSelf. The metrics are validity (V), diversity (D), and membership (M). The \textbf{better} scores are in bold.}
  \label{tab:molgen_results}
\end{table*}

\subsubsection{Class-Specific Molecule Generation}

In Table~\ref{tab:molgen_results}, we observe that applying our proposed SmiSelf method results in improvements across all metrics. This outcome can be attributed to the one-to-many nature of the task (learn the distribution of a class from a few examples and sample from it to generate multiple new molecules), and the results indicate that the molecules decoded from the corrected SMILES successfully capture the specifics of the monomer class. Notably, while standard prompting results in very low validity for acrylate molecules generated by LLMs, these molecules—once corrected by our SmiSelf method—achieve 100\% validity and show significant improvement in the Membership metric. These findings suggest that although LLMs face challenges in generating valid SMILES strings, they can still capture class-specific molecular characteristics through low-shot examples. Furthermore, this highlights that our proposed SmiSelf method not only corrects invalid molecules but also preserves their molecular characteristics.

We also observe that, compared to standard prompting, grammar prompting does not consistently improve validity or other performance metrics. This suggests that explicitly incorporating generic SMILES grammar into the prompt may not provide additional benefits. Moreover, while the baseline method DEG achieves 100\% validity in its generated molecules, its Membership metric across all three molecular classes is lower compared to the prompting-based methods and significantly lower than that of the molecules corrected using our SmiSelf method. This is because LLMs have encountered SMILES strings during pretraining, allowing them to acquire extensive domain knowledge about molecules. In contrast, DEG cannot incorporate external knowledge beyond the 11 or 32 molecules provided in its training data. Additionally, the high computational complexity of grammar construction limits DEG to being applied only to structurally similar low-shot molecules. Results for more baselines are in Appendix~\ref{appendix:more_baseline_results}.

\section{Related Work}
\label{section:methods_validity}

\begin{table*}[hbtp]
  \centering
  \tabcolsep=4.2pt
  \resizebox{1\textwidth}{!}{
  \begin{tabular}{c|c|c|c|c|c|c|c|c|c}
    \toprule[2pt]
    Model & BLEU$\uparrow$  & EM$\uparrow$ & Levenshtein$\downarrow$ & MACCS FTS$\uparrow$ & RDK FTS$\uparrow$ & Morgan FTS$\uparrow$ & FCD$\downarrow$ & Text2Mol$\uparrow$ & Validity$\uparrow$ \\
    \midrule
    Ground Truth & 1.000 & 1.000 & 0.000 & 1.000 & 1.000 & 1.000 & 0.00 & 0.609 & 1.000 \\
    \midrule
    \rowcolor[HTML]{EFEFEF}
    \multicolumn{10}{c}{\textbf{Constrained Decoding for Generation-Time Correction}}\\
    \midrule
    MolT5-Large & 0.858 & 0.318 & 15.957 & 0.890 & 0.813 & 0.750 & 0.38 & 0.590 & 0.958 \\
    MolT5-Large-HV & 0.810 & 0.314 & 16.758 & 0.872 & 0.786 & 0.722 & 0.44 & 0.582 & 0.996 \\
    &  \highlight{LightPink}{-5.59\%} & \highlight{LightPink}{-1.26\%} & \highlight{LightPink}{+5.02\%} & 	\highlight{LightPink}{-2.02\%} & \highlight{LightPink}{-3.32\%} & \highlight{LightPink}{-3.73\%} & \highlight{LightPink}{+15.79\%} & \highlight{LightPink}{-1.36\%} & \highlight{LightCyan}{+3.97\%} \\
    \midrule
    \rowcolor[HTML]{EFEFEF}
    \multicolumn{10}{c}{\textbf{Training Generative Models for Post-Hoc Correction}}\\
    \midrule
    $\text{TGM-DLM}_{\text{w/o\ corr}}$& 0.828 & 0.242 & 16.897 & 0.874 & 0.771 & 0.722 & 0.89 & 0.589 & 0.789 \\
    TGM-DLM & 0.826 & 0.242 & 17.003 & 0.854 & 0.739 & 0.688 & 0.77 & 0.581 & 0.871 \\
    &  \highlight{LightPink}{-0.24\%} & 0.00\% & \highlight{LightPink}{+0.63\%} & 	\highlight{LightPink}{-2.29\%} & \highlight{LightPink}{-4.15\%} & \highlight{LightPink}{-4.71\%} & \highlight{LightCyan}{-13.48\%} & \highlight{LightPink}{-1.36\%} & \highlight{LightCyan}{+10.39\%} \\
    \midrule
    \rowcolor[HTML]{EFEFEF}
    \multicolumn{10}{c}{\textbf{Our Methods for Post-Hoc Correction}}\\
    \midrule
    GPT-4 (10-shot) & 0.857 & 0.280 & 17.14 & 0.903 & 0.805 & 0.739 & 0.41 & 0.593 & 0.899 \\
    + Invalid SELFIES Editing & 0.682 & 0.179 & 26.596 & 0.756 & 0.624 & 0.541 & 1.666 & 0.468 & 1.000 \\
    & \highlight{LightPink}{-20.42\%} & \highlight{LightPink}{-36.07\%} & \highlight{LightPink}{+55.17\%} & \highlight{LightPink}{-16.28\%} & \highlight{LightPink}{-22.48\%} & \highlight{LightPink}{-26.79\%} & \highlight{LightPink}{+306.34\%} & \highlight{LightPink}{-21.08\%} & \highlight{LightCyan}{+11.23\%} \\
    \midrule
    GPT-4 (10-shot) & 0.857 & 0.280 & 17.14 & 0.903 & 0.805 & 0.739 & 0.41 & 0.593 & 0.899 \\
    + LLM Corrector (GPT-3.5) & 0.772 & 0.280 & 22.220 & 0.890 &  0.784 & 0.710 &  0.375 & 0.582 & 0.981 \\
    & \highlight{LightPink}{-9.92\%} & 0.00\% & \highlight{LightPink}{+29.64\%} & \highlight{LightPink}{-1.44\%} & \highlight{LightPink}{-2.61\%} & \highlight{LightPink}{-3.92\%} & \highlight{LightCyan}{-8.54\%} & \highlight{LightPink}{-1.85\%} & \highlight{LightCyan}{+9.12\%} \\
    \midrule
    GPT-4 (10-shot) & 0.857 & 0.280 & 17.14 & 0.903 & 0.805 & 0.739 & 0.41 & 0.593 & 0.899 \\
    + SmiSelf & 0.846 & 0.282 & 17.668 & 0.892 &  0.789 & 0.718 &  0.312 & 0.584 & 1.000 \\
    & \highlight{LightPink}{-1.28\%} & \highlight{LightCyan}{+0.71\%} & \highlight{LightPink}{+3.08\%} & \highlight{LightPink}{-1.22\%} & \highlight{LightPink}{-1.99\%} & \highlight{LightPink}{-2.84\%} & \highlight{LightCyan}{-23.90\%} & \highlight{LightPink}{-1.52\%} & \highlight{LightCyan}{+11.23\%} \\
    \bottomrule[1.2pt]
  \end{tabular}%
  }
  \caption{Results of methods for improving the validity of text-based molecule generation, with relative improvements marked in blue and declines marked in pink.}
  \label{tab:validity-comparison}%
\end{table*}%

In this section, we introduce various methods to improve the validity of generated molecules. For a broader discussion, see Appendix \ref{appendix:broader-related-work}.

The existing potential solutions for generating valid SMILES with LLMs can be categorized into training-time correction, generation-time correction, and post-hoc correction \cite{pan2024automatically}. Since training-time correction is limited by the infeasibility of fine-tuning giant closed-source LLMs, we will focus on generation-time correction and post-hoc correction.

\noindent\textbf{Constrained Decoding for Generation-Time Correction.} Constrained decoding is a technique used to enforce constraints on language model outputs. It restricts model outputs to adhere to predefined constraints without requiring retraining or modifications to the model architecture \cite{geng2023grammar, geng2024sketch}. While constrained decoding can improve the validity of molecule generation, it reduces the search space and significantly lowers other metrics \cite{wang2024grammar, edwards2022translation}. Additionally, constrained decoding increases the number of LLM API calls.

\noindent\textbf{Training Generative Models for Post-Hoc Correction.} Another possible approach is training generative models to correct invalid SMILES generated by LLMs post hoc. Theoretically, invalid SMILES strings could also be corrected using translator models, as employed in the field of grammatical error correction \cite{yuan2016grammatical}. However, this approach requires both invalid and ground-truth molecules to form input-output pairs for training, and thus may be task-specific \cite{gong2024text, zheng2019predicting}. Moreover, such models cannot correct 100\% of invalid outputs, and the percentage of corrected outputs varies across different invalid output generators \cite{schoenmaker2023uncorrupt}.

To compare our methods with these approaches, we calculate the relative improvement in text-based molecule generation. As shown in Table~\ref{tab:validity-comparison}, all methods come with trade-offs. SmiSelf provides a promising approach for generating 100\% valid molecules with LLMs, while keeping the performance on other metrics.

\section{Conclusion}

This paper studies how to ensure that the molecules generated by LLMs are 100\% valid. To this end, we first propose Invalid SELFIES Editing and LLMs as post-hoc correctors. Through our experiments, we find that: 1) LLMs perform worse when using SELFIES compared to SMILES; 2) LLMs face challenges in correcting and refining the invalid SMILES they generate. We then present SmiSelf, a cross-chemical language framework for invalid SMILES correction. We propose converting invalid SMILES generated by LLMs into SELFIES and transforming them back into SMILES, leveraging the mechanism of SELFIES to correct the SMILES. Experiments demonstrate that SmiSelf effectively corrects invalid SMILES generated by LLMs, ensuring 100\% validity while preserving their original molecular characteristics and maintaining or even enhancing performance on other metrics. SmiSelf helps expand the practical applications of LLMs in the biomedical domain and is compatible with all SMILES-based generative models.

\section*{Limitations}

Like other post-hoc correction methods, SmiSelf introduces some distortion in the correction process for the text-based molecule generation task, which can lead to corrected molecules deviating further from the ground truth and being less aligned with the given descriptions.

\section*{Acknowledgements}

We thank Kai-Wei Chang from the UCLA NLP group for the support and suggestions. This work is supported by the National Key R\&D Program of China under Grant No.\ 2024YFA1012700 and No.\ 2023YFF0725100, by the National Natural Science Foundation of China (NSFC) under Grant No.\ 62372159, No.\ 62402410, and No.\ U22B2060, by Guangdong Provincial Project (No.\ 2023QN10X025), by Guangdong Basic and Applied Basic Research Foundation under Grant No.\ 2023A1515110131, by Guangzhou Municipal Science and Technology Bureau under Grant No.\ 2024A04J4454, by Guangzhou Municipal Education Bureau (No.\ 2024312263), by Guangzhou Industrial Information and Intelligent Key Laboratory Project (No.\ 2024A03J0628), by Guangzhou Municipal Key Laboratory of Financial Technology Cutting-Edge Research (No.\ 2024A03J0630), by NTU Start-Up Grant, and by the Ministry of Education, Singapore, under its Academic Research Fund Tier 1 (RG22/24) and Academic Research Fund Tier 2 (FY2025) (Grant MOE-T2EP20124-0009).

\bibliography{custom}

\appendix

\section{Broader Related Work}
\label{appendix:broader-related-work}

\textbf{Molecule Generation.} Existing methods can be categorized according to their molecular representation. Molecules, for example, can be treated as chemical graphs \cite{luo2021graphdf, vignacdigress}, combinations of substructures \cite{jin2018junction}, or three-dimensional objects \cite{hoogeboom2022equivariant, xu2023geometric}. However, these approaches have yet to consistently surpass the earlier chemical language models \cite{gomez2018automatic, flam2022language}. These models represent molecules as text strings, typically using the SMILES \cite{weininger1988smiles} or SELFIES \cite{krenn2020self} formats.

\noindent\textbf{Validity of Generated Molecules.} SMILES \cite{weininger1988smiles} strings have been a prominent molecular representation since they were invented. However, the SMILES representation is not inherently robust, meaning that generative models are likely to produce strings that do not represent valid molecules. A large body of work has been dedicated to addressing this issue in recent years, whether by developing alternative textual representations of molecules \cite{o2018deepsmiles, krenn2020self, cheng2023group}, methods that generate valid SMILES by design \cite{kusner2017grammar, dai2018syntax}, or techniques to correct invalid SMILES post hoc \cite{schoenmaker2023uncorrupt, zheng2019predicting, kim2024data, gong2024text}.

\section{Molecule-Caption Generation}
\label{appendix:molecule-caption-generation}

\subsection{Evaluation Metrics}

\textbf{BLEU (Bilingual Evaluation Understudy)} measures the similarity between generated and reference texts (e.g., molecule captions). Higher is better.

\noindent\textbf{ROUGE (Recall-Oriented Understudy for Gisting Evaluation)} measures overlap between generated and reference molecule captions. Higher is better.

\noindent\textbf{METEOR (Metric for Evaluation of Translation with Explicit ORdering)} measures similarity between generated and reference molecule captions, considering precision, recall, synonyms, and word order. Higher is better.

\noindent\textbf{EM (Exact Match)} checks if the generated molecule exactly matches the ground truth. Higher is better.

\noindent\textbf{Levenshtein (Edit Distance)} measures the minimum number of insertions, deletions, or substitutions needed to convert one string to another. Lower is better.

\noindent\textbf{MACCS FTS (MACCS Fingerprint Tanimoto Similarity)} measures Tanimoto similarity between target and generated molecules using MACCS fingerprints \cite{durant2002reoptimization}. Higher is better.

\noindent\textbf{RDK FTS (RDKit Fingerprint Tanimoto Similarity)} is similar to MACCS FTS but uses RDKit fingerprints \cite{schneider2015get}. Higher is better.

\noindent\textbf{Morgan FTS (Morgan Fingerprint Tanimoto Similarity)} measures Tanimoto similarity of target and generated molecules using Morgan fingerprints \cite{rogers2010extended}. Higher is better.

\noindent\textbf{FCD (Fréchet ChemNet Distance)} measures the distance between generated and target molecule distributions using ChemNet \cite{mayr2018large}. Lower is better.

\noindent\textbf{Text2Mol} measures relevance between textual descriptions and generated molecules using the Text2Mol model \cite{edwards2021text2mol}. Higher is better.

\noindent\textbf{Validity} measures whether generated strings are valid chemical representations using RDKit \cite{landrum2013rdkit}. Higher is better.

\subsection{Datasets}

We utilize the ChEBI-20 dataset \cite{edwards2021text2mol}, which contains 33,010 molecule-caption pairs. The dataset is split into 80\% for training, 10\% for validation, and 10\% for testing. For in-context learning, the training set serves as a local database to retrieve n-shot examples.

\begin{table*}[htb]
  \center
  \small
  \begin{tabular}{ l c cccc c cccc c cccc  }
    \toprule
    & \,\, & \multicolumn{3}{c}{\textbf{Acrylates}} & \,\, & \multicolumn{3}{c}{\textbf{Chain Extenders}} &\,\,  &\multicolumn{3}{c}{ \textbf{Isocyanates} }\\
    \hspace{4mm}Model &  & V & D & M  & & V & D & M & & V & D & M \\
    \midrule
    \multicolumn{13}{l}{\textit{Task-Specific}}\\
    \hspace{4mm}JT-VAE & & 100  & 0.29  & 49  & & 100  & 0.62 & 80  & & 100  & 0.72  & 67 \\
    \hspace{4mm}HierVAE & & 100  & 0.83  & 1  & & 100  & 0.83 & 44  & & 100  & 0.83  & 0 \\
    \hspace{4mm}MHG & & 100  & 0.89  & 1  & & 100  & 0.90 & 41  & & 100  & 0.88  & 12 \\
    \hspace{4mm}DEG & & 100  & 0.83  & 30  & & 100  & 0.86 & 98  & & 100  & 0.93  & 83 \\
    \midrule
    \multicolumn{13}{l}{\textit{Prompting + SmiSelf}}\\
    \hspace{4mm}GPT-3.5 & & 100  & 0.75  & 83  & & 100  & 0.81 & 99  & & 100  & 0.82  & 100 \\
    \bottomrule
  \end{tabular}
  \caption{Results for class-specific molecule generation. The metrics are validity (V), diversity (D), and membership (M). Higher is better for all metrics.}
  \label{tab:more_baseline_results}
\end{table*}

\subsection{Baselines}

\textbf{RNN.} RNN-GRU \cite{cho2014learning} with a 4-layer bidirectional encoder, trained from scratch on ChEBI-20.

\noindent\textbf{Transformer.} A vanilla Transformer \cite{vaswani2017attention} with six encoder–decoder layers, trained from scratch on ChEBI-20.

\noindent\textbf{T5.} A model based on T5 \cite{raffel2020exploring}, pre-trained on C4 and directly fine-tuned on ChEBI-20, with small, base, and large variants. No molecular knowledge is used in pre-training.

\noindent\textbf{MolT5.} MolT5 \cite{edwards2022translation} is initialized from pre-trained T5, jointly pre-trained on ZINC-15 SMILES \cite{sterling2015zinc} and C4 text \cite{raffel2020exploring}, and then fine-tuned on ChEBI-20. It is available in small, base, and large sizes.

\noindent\textbf{LLMs.} GPT-3.5 (GPT-3.5-Turbo), GPT-4 (GPT-4-0314) \cite{achiam2023gpt}, and GPT-4o-mini \cite{hurst2024gpt} are accessed via the OpenAI API. The open-source LLMs LLaMA2-7B \cite{touvron2023llama} and LLaMA3-8B \cite{grattafiori2024llama} are used without fine-tuning. Inputs follow the five-part structure of \cite{li2024empowering}: role, task, examples, output instruction, and user prompt, with examples retrieved using BM25 \cite{robertson2009probabilistic} (text-based molecule generation) or Morgan Fingerprint \cite{butina1999unsupervised} similarity (molecule captioning).

\section{Class-Specific Molecule Generation}
\label{appendix:more_baseline_results}

We compare our method with four baselines for class-specific molecule generation: JT-VAE \cite{jin2018junction}, HierVAE \cite{jin2020hierarchical}, MHG \cite{kajino2019molecular}, and DEG \cite{guo2022data}, as shown in Table~\ref{tab:more_baseline_results}.

From the results, we observe that the vocabulary-based method JT-VAE fails to extract a vocabulary that enables it to generate diverse molecules on small datasets. HierVAE, another vocabulary-based method with a more diverse vocabulary, addresses this limitation; however, its low membership scores indicate that it does not capture class-specific characteristics. Among grammar-based methods, MHG employs fine-grained rules that simply attach atoms, resulting in high diversity. Nevertheless, these rules fail to capture domain-specific characteristics when compared to another grammar-based method, DEG.

Overall, the results demonstrate that molecules generated using the prompting-based method and subsequently corrected with our proposed SmiSelf successfully capture class-specific features and consistently achieve stable performance. These findings clearly distinguish our approach from the baselines.

\begin{table*}[t]
  \centering
  \small
  \begin{adjustbox}{width=\linewidth}
    \begin{tabular}{c|c|c|c|c|c|c|c|c|c|c|c|c|c|c}
      \toprule
      \textbf{State} & \textbf{[$\epsilon$]} & \textbf{[F]} & \textbf{[=O]} & \textbf{[\#N]} & \textbf{[O]} & \textbf{[N]} & \textbf{[=N]} & \textbf{[C]} & \textbf{[=C]} & \textbf{[\#C]} & \textbf{[Branch1]} & \textbf{[Branch2]} & \textbf{[Branch3]} & \textbf{[Ring]} \\
      \midrule
      X\textsubscript{0} & X\textsubscript{0} & F X\textsubscript{1} & O X\textsubscript{2} & N X\textsubscript{3} & O X\textsubscript{2} & N X\textsubscript{3} & N X\textsubscript{3} & C X\textsubscript{4} & C X\textsubscript{4} & C X\textsubscript{4} & ign X\textsubscript{0} & ign X\textsubscript{0} & ign X\textsubscript{0} & ign X\textsubscript{0} \\ \midrule
      X\textsubscript{1} & $\epsilon$ & F & O & N & O X\textsubscript{1} & N X\textsubscript{2} & N X\textsubscript{2} & C X\textsubscript{3} & C X\textsubscript{3} & C X\textsubscript{3} & ign X\textsubscript{1} & ign X\textsubscript{1} & ign X\textsubscript{1} & R(N) \\
      \midrule
      X\textsubscript{2} & $\epsilon$ & F & =O & =N & O X\textsubscript{1} & N X\textsubscript{2} & =N X\textsubscript{1} & C X\textsubscript{3} & =C X\textsubscript{2} & =C X\textsubscript{2} & B(N, X\textsubscript{5})X\textsubscript{1} & B(N, X\textsubscript{5})X\textsubscript{1} & B(N, X\textsubscript{5})X\textsubscript{1} & R(N) X\textsubscript{1} \\
      \midrule
      X\textsubscript{3} & $\epsilon$ & F & =O & \#N & O X\textsubscript{1} & N X\textsubscript{2} & =N X\textsubscript{1} & C X\textsubscript{3} & =C X\textsubscript{2} & \#C X\textsubscript{1} & B(N, X\textsubscript{5})X\textsubscript{2} & B(N, X\textsubscript{6})X\textsubscript{1} & B(N, X\textsubscript{5})X\textsubscript{2} & R(N) X\textsubscript{2} \\
      \midrule
      X\textsubscript{4} & $\epsilon$ & F & =O & \#N & O X\textsubscript{1} & N X\textsubscript{2} & =N X\textsubscript{1} & C X\textsubscript{3} & =C X\textsubscript{2} & \#C X\textsubscript{1} & B(N, X\textsubscript{5})X\textsubscript{3} & B(N, X\textsubscript{7})X\textsubscript{1} & B(N, X\textsubscript{6})X\textsubscript{2} & R(N) X\textsubscript{3} \\
      \midrule
      X\textsubscript{5} & C & F & O & N & O X\textsubscript{1} & N X\textsubscript{2} & N X\textsubscript{2} & C X\textsubscript{3} & C X\textsubscript{3} & C X\textsubscript{3} & X\textsubscript{5} & X\textsubscript{5} & X\textsubscript{5} & X\textsubscript{5} \\ \midrule
      X\textsubscript{6} & C & F & =O & =N & O X\textsubscript{1} & N X\textsubscript{2} & =N X\textsubscript{1} & C X\textsubscript{3} & =C X\textsubscript{2} & =C X\textsubscript{2} & X\textsubscript{6} & X\textsubscript{6} & X\textsubscript{6} & X\textsubscript{6} \\ \midrule
      X\textsubscript{7} & C & F & =O & \#N & O X\textsubscript{1} & N X\textsubscript{2} & =N X\textsubscript{1} & C X\textsubscript{3} & =C X\textsubscript{2} & \#C X\textsubscript{1} & X\textsubscript{7} & X\textsubscript{7} & X\textsubscript{7} & X\textsubscript{7} \\ \midrule
      N & 1 & 2 & 3 & 4 & 5 & 6 & 7 & 8 & 9 & 10 & 11 & 12 & 13 & 14 \\
      \bottomrule
    \end{tabular}
  \end{adjustbox}
  \caption{Derivation rules of SELFIES for small organic molecules.}
  \label{tab:SELFIES_rules}%
\end{table*}

\section{SMILES vs. SELFIES}
\label{appendix:selfies}
SMILES (Simplified Molecular-Input Line-Entry System) \cite{weininger1988smiles} is the de facto standard representation in cheminformatics. In SMILES, molecules are represented as a chain of atoms, written as letters in a string. Branches in the molecule are enclosed in parentheses, while ring closures are indicated by two matching numbers. Although the SMILES grammar is simple, it allows for the description of complex structures, as well as properties such as stereochemistry. However, SMILES lacks a mechanism to ensure that molecular strings are valid in terms of both syntax and physical principles.

SELFIES (SELF-referencIng Embedded Strings) \cite{krenn2020self}, on the other hand, is a 100\% robust molecular string representation. That is, SELFIES cannot produce an invalid molecule, as every combination of symbols in the SELFIES alphabet corresponds to a chemically valid graph. SELFIES is a formal grammar with derivation rules (Table~\ref{tab:SELFIES_rules}). It can be understood as a small computer program with minimal memory that guarantees 100\% robust derivation. The SELFIES grammar is specifically designed to eliminate both syntactically and semantically invalid molecules, which is especially important in generative tasks.

\section{Syntactic Validity vs. Semantic Validity}
\label{appendix:syntactic_semantic_validity}

\textbf{Syntactic validity} refers to whether the string conforms to specific syntactic rules and can be parsed into a molecular graph. For example, the SMILES string \texttt{C\#C=C} is syntactically valid because it adheres to SMILES syntax rules.

\noindent\textbf{Semantic validity} refers to whether the molecular graph represented by the string adheres to fundamental chemical rules, such as the valence rules for atoms. For example, the SMILES string \texttt{C\#C=C} is semantically invalid because the middle carbon (bonded via \texttt{\#} and \texttt{=}) exceeds carbon’s maximum valency of 4.

A syntactically invalid string is always semantically invalid because it cannot be parsed into a molecular graph and therefore cannot be assessed for semantic validity.

We provide examples of three possible cases:

\begin{itemize}
\item \textit{Syntactically invalid:} The SMILES string \texttt{C\#C=C)} is syntactically invalid because of the non-matched \texttt{)}.

\item \textit{Syntactically valid but semantically invalid:} The SMILES string \texttt{C\#C=C} is syntactically valid, but the middle carbon (bonded via \texttt{\#} and \texttt{=}) exceeds carbon’s maximum valency of 4, which violates chemical rules and is therefore semantically invalid.

\item \textit{Both syntactically and semantically valid:} The SMILES string \texttt{C=C=C} is both syntactically and semantically valid, representing a molecule that adheres to both syntactic and chemical rules.
\end{itemize}

\section{Fine-tuning vs. SmiSelf}
Although fine-tuning can be applied to achieve higher validity and improve other metrics, we would like to highlight several crucial factors to consider when deciding whether to use it:
\begin{itemize}
\item availability of training data

\item computational cost of fine-tuning

\item time cost of fine-tuning

\item performance improvement

\item feasibility of training LLMs
\end{itemize}
In contrast, our proposed SmiSelf:
\begin{itemize}
\item does not require training data

\item eliminates the computational cost of fine-tuning, with only a small overhead

\item is rapid

\item ensures 100\% validity while preserving molecular characteristics and maintaining or even enhancing performance on other metrics

\item is compatible with all SMILES-based generative models
\end{itemize}

\section{Prompts}
\label{appendix:prompt}

\textbf{Prompt for text-based molecule generation:}

\begin{lstlisting}[style=markdownstyle]
# System Prompt

You are now working as an excellent expert in chemisrty and drug discovery. Given the caption of a molecule, your job is to predict the SMILES representation of the molecule. The molecule caption is a sentence that describes the molecule, which mainly describes the molecule's structures, properties, and production. You can infer the molecule SMILES representation from the caption.

Example 1:
```
Instruction: Given the caption of a molecule, predict the SMILES representation of the molecule.
Input: The molecule is a steroid ester that is pregn-4-en-21-yl acetate substituted by oxo group at positions 3 and 20, a methyl group at position 6 and hydroxy groups at positions 11 and 17 respectively. It is a 3-oxo-Delta(4) steroid, a steroid ester, an 11beta-hydroxy steroid, a 17alpha-hydroxy steroid, a 20-oxo steroid and a tertiary alpha-hydroxy ketone. It derives from a hydride of a pregnane.
```

Your output should be:
```
{"molecule": "C[C@H]1C[C@H]2[C@@H]3CC[C@@]([C@]3(C[C@@H]([C@@H]2[C@@]4(C1=CC(=O)CC4)C)
O)C)(C(=O)COC(=O)C)O"}
```

Your response should only be in the exact JSON format above; THERE SHOULD BE NO OTHER CONTENT INCLUDED IN YOUR RESPONSE.

# User Prompt

Input: The molecule is a steroid ester that is methyl (17E)-pregna-4,17-dien-21-oate substituted by oxo groups at positions 3 and 11. It is a 3-oxo-Delta(4) steroid, an 11-oxo steroid, a steroid ester and a methyl ester. It derives from a hydride of a pregnane.

\end{lstlisting}

\noindent\textbf{Prompt for molecule captioning:}

\begin{lstlisting}[style=markdownstyle]
# System Prompt

You are now working as an excellent expert in chemisrty and drug discovery. Given the SMILES representation of a molecule, your job is to predict the caption of the molecule. The molecule caption is a sentence that describes the molecule, which mainly describes the molecule's structures, properties, and production.

Example 1:
```
Instruction: Given the SMILES representation of a molecule, predict the caption of the molecule.
Input: C[C@]12CCC(=O)C=C1CC[C@@H]3[C@@H]2C(=O)C[C@]4([C@H]3CCC4=O)C
```

Your output should be:
```
{"caption": "The molecule is a 3-oxo Delta(4)-steroid that is androst-4-ene carrying three oxo-substituents at positions 3, 11 and 17. It has a role as an androgen, a human urinary metabolite, a marine metabolite and an EC 1.1.1.146 (11beta-hydroxysteroid dehydrogenase) inhibitor. It is a 3-oxo-Delta(4) steroid, a 17-oxo steroid, an androstanoid and an 11-oxo steroid. It derives from a hydride of an androstane."}
```

Your response should only be in the JSON format above; THERE SHOULD BE NO OTHER CONTENT INCLUDED IN YOUR RESPONSE.

# User Prompt

Input: C[C@]12CCC(=O)C=C1CC[C@@H]3[C@@H]2C(=O)C[C@]\\4([C@H]3CC/C4=C/C(=O)OC)C

\end{lstlisting}

\noindent\textbf{Prompt for LLMs as correctors:}

\begin{lstlisting}[style=markdownstyle]
# System Prompt

You are now working as an excellent expert in chemisrty and drug discovery. Given the invalid SMILES representation and the caption of a molecule, your job is to predict the valid SMILES representation of the molecule. The molecule caption is a sentence that describes the molecule, which mainly describes the molecule's structures, properties, and production. You can infer the molecule SMILES representation from the caption.

Task Format
```
Instruction: Given the invalid SMILES representation and the caption of a molecule, predict the valid SMILES representation of the molecule.
Input:
Invalid SMILES Representation: [INVALID_SMILES_REPRESENTATION_MASK]
Caption: [CAPTION_MASK]
```

Your output should be:
```
{"molecule": "[VALID_SMILES_REPRESENTATION_MASK]"}
```

Your response should only be in the exact JSON format above; THERE SHOULD BE NO OTHER CONTENT INCLUDED IN YOUR RESPONSE.

# User Prompt

Input:
Invalid SMILES Representation: C[C@H]1[C@H]([C@H]([C@@H]([C@@H](O1)O[C@@H]2[C@H]([C@H]([C@H](O[C@H]2O)
CO)O[C@H]3[C@@H]([C@H]([C@@H]([C@H](O3)CO)O)O)NC(=O)C)O)O)NC(=O)C)O)O
Caption: The molecule is a branched amino tetrasaccharide consisting of N-acetyl-beta-D-glucosamine having two alpha-L-fucosyl residues at the 3- and 6-positions as well as an N-acetyl-beta-D-glucosaminyl residue at the 4-position. It has a role as a carbohydrate allergen. It is a glucosamine oligosaccharide and an amino tetrasaccharide. It derives from an alpha-L-Fucp-(1->3)-[alpha-L-Fucp-(1->6)]-beta-D-GlcpNAc.

\end{lstlisting}

\noindent\textbf{Prompt for class-specific molecule generation:}

\begin{lstlisting}[style=markdownstyle]
You are an expert in chemistry. You are given a list of acrylates molecules in SMILES format. You are asked to write another acrylates molecule in SMILES format.
Molecule: C=CC(=O)OCCCCCCOC(=O)C=C
Molecule: CCCCCCOC(=O)C=C
Molecule: CCCOC(=O)C(=C)C
Molecule: CCC(C)OC(=O)C(=C)C
Molecule: CCC(COCCCOC(=O)C=C)(COCCCOC(=O)C=C)COCCCOC(=O)C=C
Molecule: C=CC(=O)OC1=CC=CC=C1
Molecule: CCC(C)OC(=O)C=C
Molecule: CCCCCCCCOC(=O)C(=C)C
Molecule: C=CC(=O)OC1=C(C(=C(C(=C1F)F)F)F)F
Molecule: CC(=C)C(=O)OCCOC1=CC=CC=C1
Molecule: CCCCCCCCCCCCOC(=O)C(=C)C
Molecule: CC(=C)C(=O)OC
Molecule: C=CC(=O)OCC(CO)(COCC(COC(=O)C=C)(COC(=O)C=C)COC(=O)C=C)COC(=O)C=C
Molecule: CC(C)CCCCCCCOC(=O)C=C
Molecule: CCOCCOC(=O)C(=C)C
Molecule: C=CC(=O)OCC1=CC=CC=C1
Molecule: CCCCOC(=O)C=C
Molecule: CCC(COCC(CC)(COC(=O)C=C)COC(=O)C=C)(COC(=O)C=C)COC(=O)C=C
Molecule: CC(=C)C(=O)OCC1=CC=CC=C1
Molecule: CC1CC(CC(C1)(C)C)OC(=O)C(=C)C
Molecule: COC(=O)C=C
Molecule: CC(=C)C(=O)OC1CC2CCC1(C2(C)C)C
Molecule: CCCOC(=O)C=C
Molecule: COCCOC(=O)C=C
Molecule: C=CC(=O)OCCC1=CC=CC=C1
Molecule: C=CC(=O)OCC(COCC(COC(=O)C=C)(COC(=O)C=C)COC(=O)C=C)(COC(=O)C=C)COC(=O)C=C
Molecule: CC(=C)C(=O)OC1=CC=CC=C1
Molecule: CCCCC(CC)COC(=O)C(=C)C
Molecule: CC(C)(COCCCOC(=O)C=C)COCCCOC(=O)C=C
Molecule: C=CC(=O)OCC(CO)(COC(=O)C=C)COC(=O)C=C
Molecule: CCCCOCCOC(=O)C(=C)C
Molecule: CC(C)COC(=O)C(=C)C
Molecule:

\end{lstlisting}

\end{document}